\documentclass[10pt,twocolumn,letterpaper]{article}
\usepackage[pagenumbers]{cvpr}
\definecolor{cvprblue}{rgb}{0.21,0.49,0.74}
\definecolor{GUIcolor}{RGB}{210,237,238}
\definecolor{Poolcolor}{RGB}{252,232,182}
\definecolor{RScolor}{RGB}{248,213,208}
\definecolor{GUIdark}{RGB}{171,220,224}
\definecolor{Pooldark}{RGB}{250,220,144}
\definecolor{RSdark}{RGB}{243,176,167}
\definecolor{darkred}{rgb}{0.8, 0, 0}
\definecolor{darkgreen}{rgb}{0, 0.5, 0}
\usepackage[pagebackref,breaklinks,colorlinks,citecolor=cvprblue]{hyperref}
\usepackage{multirow}
\usepackage{pifont}
\usepackage{cuted}
\usepackage{graphicx}
\usepackage{colortbl}
\usepackage{makecell}
\makeatletter
\def\thanks#1{\protected@xdef\@thanks{\@thanks
        \protect\footnotetext{#1}}}
\makeatother

\title{
SmartAgent: Chain-of-User-Thought for Embodied Personalized Agent \\ in Cyber World 
}

\author{Jiaqi Zhang, Chen Gao, Liyuan Zhang, Yong Li\textsuperscript{*}, Hongzhi Yin\textsuperscript{*}\thanks{\textsuperscript{*}Corresponding authors: Yong Li \textless liyong07@tsinghua.edu.cn\textgreater, Hongzhi Yin \textless h.yin1@uq.edu.au\textgreater}\\
{\small Tsinghua University, University of Queensland}\\
}

\begin{document}
\maketitle
\begin{strip}
\centering
\vskip -0.45in
\centering
\includegraphics[width=0.9\textwidth]{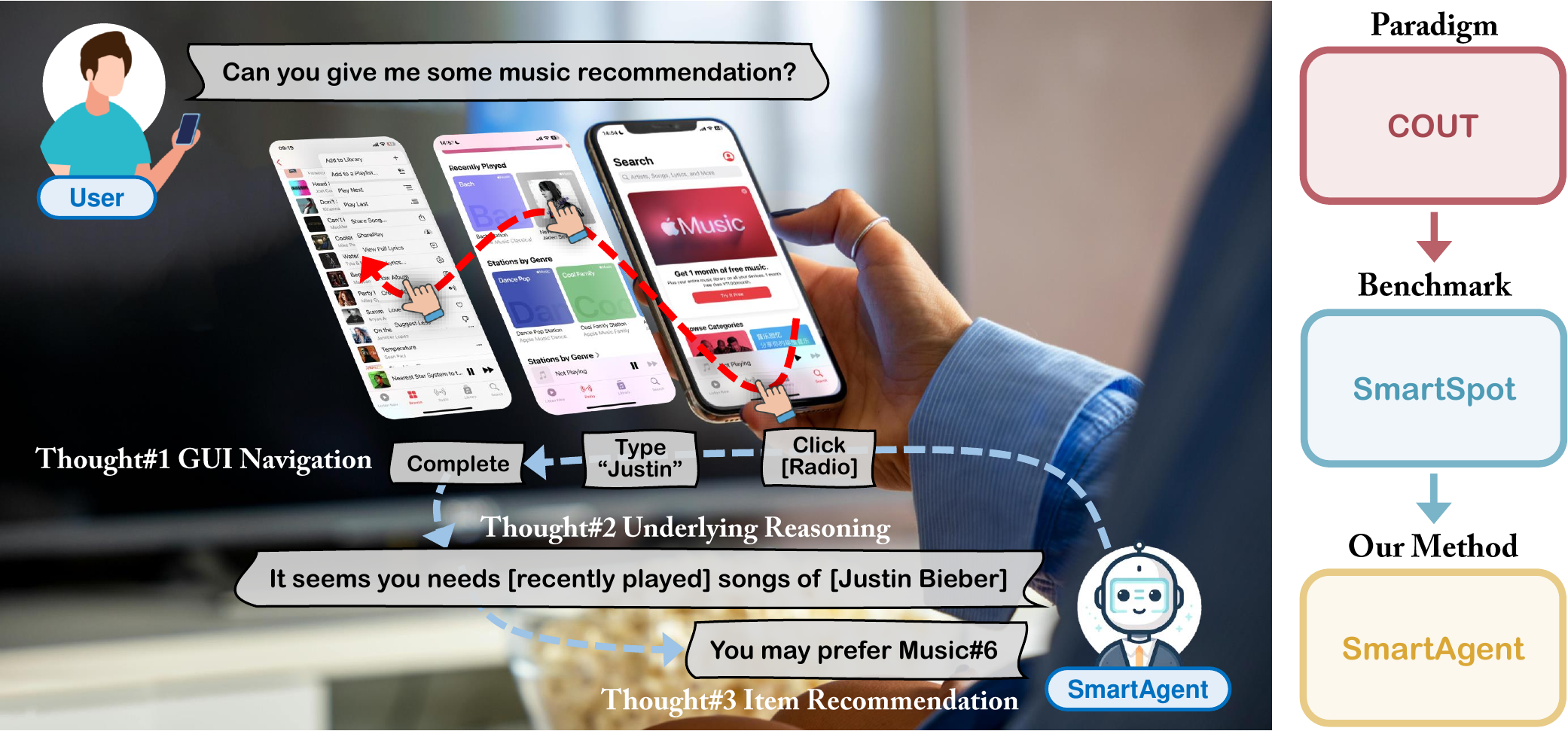}
\captionof{figure}{\textbf{The Chain-of-User-Thought (COUT) reasoning paradigm.} 
The red line shows a sequence of GUI actions, while the blue line illustrates our three-step thought process.
In Thought \#1, according to a user's instruction, an agent performs GUI actions to search for an item pool. 
In Thought \#2 with seeing the pool, the agent reasons underlying requirements behind the original instruction, as implied by the previous actions.
In Thought \#3, based on the underlying thought, the agent recommends items within the pool to complete the user's instruction.
By leveraging user-oriented thoughts, this COUT could enable full-stage embodied personalized capabilities across various information systems.
}
\label{fig:COUT}
\end{strip}
\begin{abstract}
Recent advances in embodied agents with multimodal perception and reasoning capabilities based on large vision-language models (LVLMs), excel in autonomously interacting
either real or cyber worlds, helping people make intelligent decisions in complex environments.
However, the current works are normally optimized by golden action trajectories or ideal task-oriented solutions toward a definitive goal.
This paradigm considers limited user-oriented factors, which could be the reason for their performance reduction in a wide range of personal assistant applications.
To address this, we propose \textbf{Chain-of-User-Thought (COUT)}, a novel embodied reasoning paradigm that takes a chain of thought  
from basic action thinking to explicit and implicit personalized preference thought 
to incorporate personalized factors into autonomous agent learning.
The main challenges of achieving COUT include: 
1) the definition of embodied personalized tasks,
2) the embodied environment epitomizes personalized preference, and 
3) the way to model embodied personalized actions.
To target COUT, we introduce \textbf{SmartAgent}, an agent framework perceiving cyber environments and reasoning personalized requirements as: 
1) interacting with GUI to access an item pool, 
2) generating users' explicit requirements implied by previous actions, and
3) recommending items to fulfill users' implicit requirements.
To demonstrate SmartAgent's capabilities, 
we also create a brand-new dataset \textbf{SmartSpot} that offers a full-stage personalized action-involved environment.
To our best knowledge, our work is the first to formulate the COUT process, serving as a preliminary attempt towards embodied personalized agent learning. 
Our extensive experiments on SmartSpot illuminate SmartAgent’s functionality among a series of embodied and personalized sub-tasks. 
We will release code and data upon paper notification at \url{https://github.com/tsinghua-fib-lab/SmartAgent}.
\end{abstract}    
\vspace{-0.20in}
\section{Introduction}
\label{sec:intro}
Embodied artificial intelligence~\cite{duan2022survey} is considered as a crucial stride toward Artificial General Intelligence (AGI)~\cite{duenez2023social}. 
Powered by the recent advances in large multi-modal models, embodied agents have been built upon to behave like real humans, capable of perceiving, reasoning, and acting with their surroundings in both real and cyber worlds. 
The enthusiasm for deploying such humanoid capabilities is evident in various tasks, including autonomic robotics~\cite{driess2023palm, barreiros2022haptic}, game AI~\cite{yang2025octopus, tan2024towards}, smart device assistants~\cite{rawles2024AndroidWorld, hong2023CogAgent}, and smart city~\cite{gao2024simulating, xu2023urban}.
Many of these scenarios require embodied agents to do more than follow instructions and execute actions like emotionless robots; otherwise, they are expected to serve as personal assistants attuned to human preferences in the meantime. 
For instance, for smart device assistance, agents struggle to personalize responses to ambiguous user queries such as providing music recommendations, though they fully understand the operation logic of a music player. 
Generally speaking, a fully functional embodied agent necessitates personalized perceptual capabilities, thereby enabling a comprehensive agent-environment-user triadic perception of the world
, similar to JARVIS\footnote{Stands for 'Just a Rather Very Intelligent System', a fictional artificial intelligent character in the Marvel Comics}, a fictional artificial assistant created by Iron Man
which can make various personalized intelligent decisions for the user.

However, this personalized consideration is absent among the current embodied agent works, where the optimization normally relies on golden action trajectories~\cite{rawles2024androidinthewild, deng2024mind2web} or ideal task-oriented solutions~\cite{drouin2024WorkArena, kim2024RCIPrompt}. 
Although these fixed paths can effectively accomplish task goals, 
they can only train embodied agents to be rigid task-oriented problem solvers, overlooking the multiple valid approaches that often exist as user-oriented indicators. 
Furthermore, the practical environment may exhibit unpredictable behavior, such as when new functions are involved or unseen scenes are observed.
In these cases, task-oriented agents are not flexible enough to even capture dynamic changes in basic task goals~\cite{kim2022limits, srivastava2022behavior}, let alone discern changes in user preference.
As a result, such training paradigms actually restrict the learning of user-oriented perceptual capabilities, which could be the reason for the lower performance in many embodied personalized scenarios.
For example, in the daily usage of smart devices as shown in Figure~\ref{fig:COUT}, there are no golden line actions to access a music pool but rather diverse paths imply different user intents~\cite{rawles2024AndroidWorld}.
That is, these normal embodied agents are designed to only be able to use the product (music App on mobile phones), which are not able to think about users' music preferences or extract the user intentions behind the usage habits and collected behaviors.

In this work, we propose a novel reasoning paradigm \textbf{Chain-of-User-Thought (COUT)}, which \textit{extends embodied agents from task-only optimization to personalization-oriented optimization}.
We summarize COUT process as: training agents with
\textbf{progressive thinking from basic embodied behaviors, gradually to explicit requirements reasoning, and finally to high-level implicit personalization understanding}, as shown in Figure~\ref{fig:COUT}.  
Note that the chain of user thoughts is a general paradigm and thus can be flexibly adapted to various cyber environments and enhance personalized reasoning. 
However, there exist several critical challenges to achieving COUT, as follows:
\begin{itemize}[leftmargin=*]
    \item First, the learning task of embodied personalized agents has not been systematically defined, because the task goals are often ambiguous queries from users, which go beyond the existing works that take explicit task instructions.
    \item Second, there is a lack of suitable datasets and benchmarks to support research on COUT. The commonly used datasets generally do not include personalized features.
    \item Third, the modeling of personalized features is underexplored, largely due to the absence of clear task definition and supportive training environments.
    % \vspace{0.02cm}
\end{itemize}
To support COUT research, we collect and construct the first embodied AI benchmark with explicit personalization-related evaluations, named \textbf{SmartSpot}.
It comprises five single channels and two multi-channel scenarios to simulate complex real-world environments, featuring a total of 144 episodes and over 1,400 steps.
To address the above challenges, we design \textbf{SmartAgent}, the first embodied personalized agent.
It takes visual observations of GUI screenshots and textual instructions as task input and generates multi-step thoughts. 
Specifically, the SmartAgent undergoes a two-stage training process: \textit{embodiment stage} and \textit{personalization stage}, as illustrated in Figure~\ref{fig:Two_stage_model}.
In embodiment stage, the agent takes GUI actions and item pool screenshots as visual inputs, along with other textual contexts. After a series of encoding processes, these multimodal tokens are fed into a \textit{Perceiver} model to generate specific GUI actions, regarded as Thought \#1. Based on these initial GUI thoughts, a \textit{Reasoner} model then serves to infer the Thought \#2, identifying the user's potential underlying requirements in a short textual output.
Finally in the personalization stage, powered by the deeper-level Thought \#3, the same \textit{Perceiver} directly outputs the recommendation result as ``Yes'' or ``No''. 
\begin{figure*}[t!]        
    \center
    {\includegraphics[width=\textwidth]{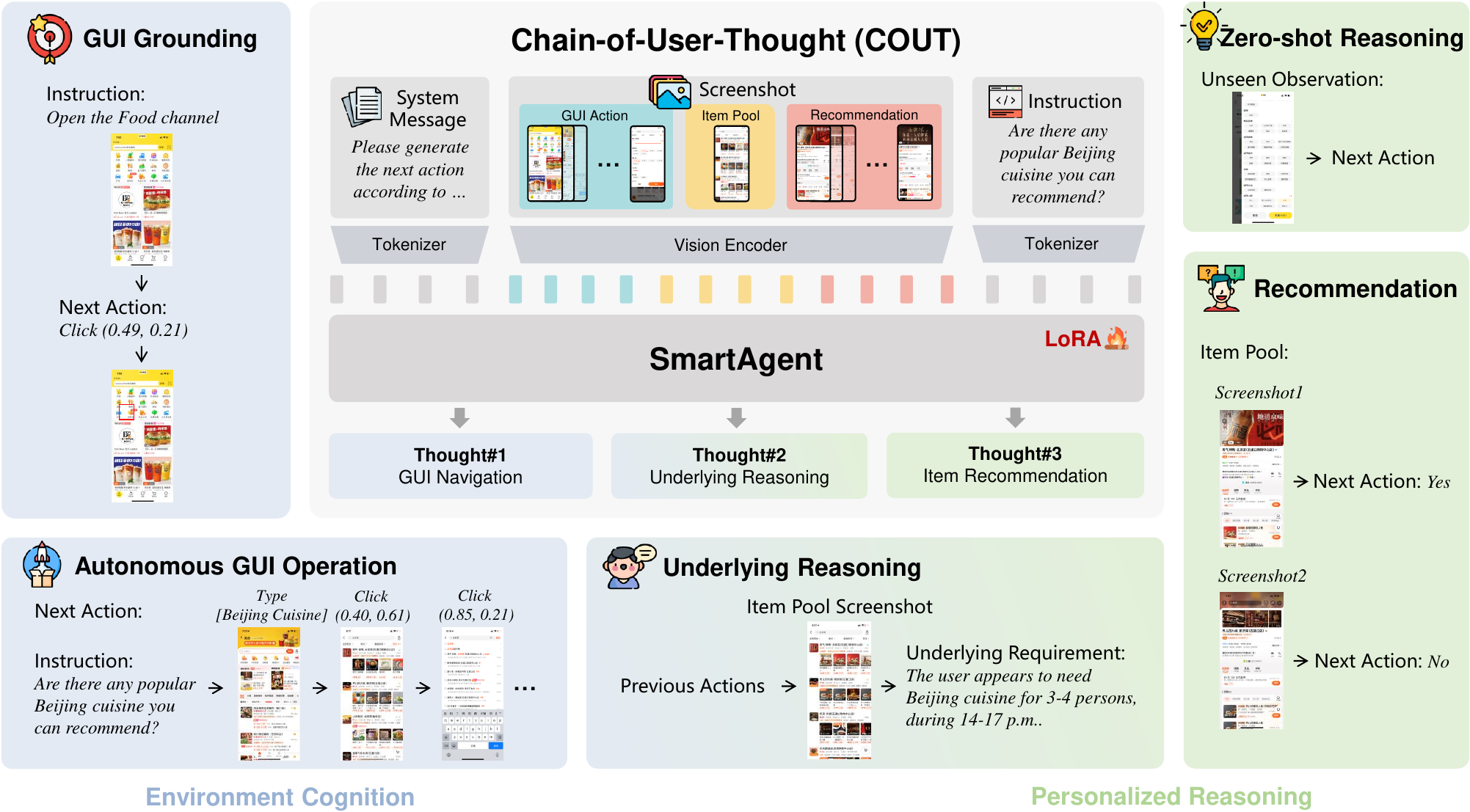}}
    \caption{
    \textcolor{black}{The full-stage embodied personalized capabilities of our proposed SmartAgent, range from basic environment cognition to advanced user personal intention reasoning.} 
    }      
\label{fig:SmartAgent}
\end{figure*}

We quantitatively evaluate SmartAgent's functionality in comprehensive embodied and personalized sub-tasks, including GUI Grounding, Autonomous GUI Operation, Underlying Reasoning, Personalized Recommendation, and Zero-shot Reasoning, as illustrated in Figure~\ref{fig:SmartAgent}. 
The results indicate 
(i) through three-step of COUT process with efficient LoRA tuning, SmartAgent achieves the first full-stage embodied personalized reasoning; 
(ii) SmartAgent delivers comparable performances to state-of-the-art task-specific models on fundamental GUI Grounding and advanced Autonomous GUI Operation tasks, capable of generating accurate action commands; 
(iii) SmartAgent shows proficiency in reasoning explicit underlying intentions, effectively bridging surface-level operations to more implicit-level user needs;
(iv) SmartAgent excels at leveraging these preceding thoughts to uncover users' implicit items requirements; 
(v) SmartAgent manifests zero-shot reasoning capabilities across new channels, a hallmark of a well-established embodied agent.
We also present qualitative evaluations to illustrate SmartAgent’s effectiveness and proficiency in personal assistant scenarios.
The contributions of this work can be summarized as follows:
\begin{itemize}[leftmargin=*]
\item We take a pioneering step to formulate a novel reasoning paradigm COUT for embodied AI, which introduces personalized concerns that have not been addressed in previous embodied AI research.
To support COUT, we construct the SmartSpot benchmark as the first case environment for further research.

\item We develop the first embodied personalized agent SmartAgent, built upon a meticulous two-stage training process that effectively implements the three-step COUT thinking process.
SmartAgent is capable of performing a range of tasks, from basic embodied functions to both explicit and implicit-level personalized reasoning.

\item Our rigorous experiments demonstrate the outstanding performance of SmartAgent on both overall evaluations and sub-tasks. SmartAgent demonstrates notable embodied personalized reasoning and zero-shot reasoning capabilities.
\end{itemize}

\section{Related Work}
\label{sec:related}

\textbf{Embodied AI with Large Vision-Language Models.} 
Embodied AI marks a significant transformation from traditional artificial intelligence which relies on static datasets, to autonomous agents that learn through interactions with their environments.
Thanks to the rapid development of LVLM, embodied agents can now process pure visual and textual observation as input. 
The current effort for training such embodied agents can be categorized into main directions, considering fixed and dynamic environments respectively~\cite{rawles2024AndroidWorld}. 
Many existing approaches follow fixed environments, by comparing agents' action trajectories to pre-collected human demonstrations~\cite{huang2023embodied, yan2023gpt, mialon2023gaia, li2020mapping, zhang2024android, zhang2024AutoGUI}. 
LEO~\cite{huang2023embodied} presents a generalist agent capable of interacting with 3D environments for various vision-language tasks, following ground-truth solutions.
GAIA~\cite{mialon2023gaia} introduces a fixed dataset that evaluates agents' ability to interact with live web environments.
Towards more realistic evaluation, emerging research has introduced dynamic environments where agents learn interactively from mistakes to probe the boundaries of systems~\cite{barreiros2022haptic, he2024webvoyager, rawles2024AndroidWorld, tan2024towards, gao2024embodiedcity, xie2024osworld}. 
CRADLE~\cite{tan2024towards} tests a General Computer Control setting in real-world AAA games, serving as a milestone towards AGI in the digital world.
EmbodiedCity~\cite{gao2024embodiedcity} introduces the first platform with a high-quality 3D real environment based on a real city, providing systematic benchmark tasks for embodied intelligence.

Though the above training environments inspire promising embodied capabilities in various tasks, they primarily optimize agents as ideal problem solvers.
This approach is still unsuitable for many personal assistant scenarios where user-oriented considerations are critical for embodied decision-making. 
SmartAgent is the first that aims to address this by integrating both embodied and personalized capabilities.

\noindent \textbf{GUI Navigation.} 
Automatically execution of user instructions on smart devices, like smartphones and desktops is an advanced task, as it requires agents to perceive, reason, and finally make decisions, i.e. Graphic User Interface (GUI) Navigation. 
As a result, GUI navigation is one of the typical scenarios where personalized intention under instructions can greatly influence embodied behavior. 
Early research in this field usually relies on some intermediate observation such as GUI layout annotation or text-only HTML format~\cite{wang2023enabling, zheng2023synapse, kim2024RCIPrompt}. Recently, A number of works have utilized LVLMs’ multimodal knowledge and emergent zero-shot embodied abilities to perform GUI Navigation~\cite{zhang2024AutoGUI, yang2023Appagent, deng2024mobilebench, wang2024Mobileagent, hong2023CogAgent, ma2024cocoagent, cheng2024seeclick}. SeeClick~\cite{cheng2024seeclick} presents an LVLM-based agent that purely relies on interface screenshots, enhanced with GUI grounding-specific pre-training. COAT~\cite{zhang2024android} introduces a dataset AITZ with semantic prompts, allowing agents with fine-grained step-wise reasoning. They also demonstrate effectiveness on three classic benchmarks, AiTW~\cite{rawles2024androidinthewild}, MiniWob~\cite{shi2017world}, and Mind2Web~\cite{deng2024mind2web}. 
Despite the advance of the above premier GUI Navigators, 
they are primarily designed to respond to instructions with clear ground-truth outcomes that can be accurately evaluated through screenshots.
In practice, however, user queries often appear as ambiguous instructions, a hallmark of personal assistant tasks that previous works have not addressed.
In this paper, we take the GUI Navigation task as a case embodied environment 
to demonstrate SmartAgent's effectiveness in analyzing ambiguous user thoughts.

\textbf{Personalized Recommendation.} 
Recommendation, a classic task in the field of data mining and information retrieval, specializes in modeling user preferences based on their previous interactions. 
Recommendation models have found wide-ranging applications, such as in e-commerce websites~\cite{hou2024bridging}, search engines~\cite{moskalenko2020scalable}, streaming services~\cite{zhang2024NineRec}, and advertising systems~\cite{guo2017deepfm}. 
These applications inherently integrate closely with personal smart devices as interfaces, allowing them to collect changes in user historical interactions~\cite{yin2024device, yuan2023federated}.
Existing studies typically represent this personalized preference through sequences of item IDs~\cite{koren2009matrix, rendle2012bpr, kang2018self, yuan2019simple} or advanced item content~\cite{yuan2023go, zhang2024NineRec, Fu2024Adapter, cheng2024image}. 
With the enhancement of LVLMs, modeling user multimodal behavior has emerged as a new research hotspot in this field~\cite{li2023exploring, wei2024llmrec, geng2022recommendation, wu2024survey}.
This multimodal capability allows personalized recommendations at more diverse levels. The one includes the screen operation level's behavior, which was unobservable in the back-end item sequences before. 
In this paper, we integrate the setting of multimodal recommendation task into an embodied GUI case to enable the principles of COUT, creating the first embodied environment with explicit personalized evaluation.
\section{Chain-of-User-Thought (COUT)}
\label{sec:COUT}

\subsection{Definition}
Chain-of-User-Thought (COUT) is a reasoning paradigm where an agent controls smart devices based on both task goals and user personalized preference as follows: 
\begin{align}
\label{equ:COUT}
   \mathcal{A_\text{COUT}} = \{Action_i \mid \text{\textit{Task goal}}, \text{\textit{User preference}}\},
\end{align}
where $\mathcal{A_\text{COUT}}$ is the action space, which differs from the existing work that relies solely on task goal as:
\begin{align}
   \mathcal{A_\text{existing work}} = \{Action_i \mid \text{\textit{Task goal}}\}.
\end{align} 
Specifically, an agent is required to generate $Action_i$ through a progressive reasoning chain from \textit{basic embodied behavior level}, gradually to \textit{deeper explicit personalized reasoning level}, and finally to \textit{high implicit personalized reasoning level}.
This process requires embodied agents the ability of a multi-faceted understanding of user personalized preferences.

\begin{table*}[t!]
\caption{Datasets statistics. \textcolor{black}{Instruction Mean and Underlying Mean denote the mean length of user instructions and corresponding underlying requirements.}}
\label{tab:Datasets statistics}
\centering
\begin{tabular}{p{2.2cm}<{\centering} p{1.8cm}<{\centering} p{1.8cm}<{\centering} p{1.2cm}<{\centering} p{1.5cm}<{\centering} p{1.2cm}<{\centering} p{2.0cm}<{\centering} p{2.0cm}<{\centering}}
\toprule
 \multirow{2.5}{*}{Scenario} &\multirow{2.5}{*}{Channel} &\multirow{2.5}{*}{\#Episodes} &\multicolumn{3}{c}{\#Steps} &\multirow{2.5}{*}{\makecell[c]{Instruction\\Mean}} &\multirow{2.5}{*}{\makecell[c]{Underlying\\Mean}}  \\ 
\cmidrule(r){4-6}   
& & &\cellcolor{GUIcolor}GUI &\cellcolor{Poolcolor}Item Pool &\cellcolor{RScolor}RS & &  \\ 
\midrule
\multirow{5}{*}{Single-channel}&FOOD     &20  &7.00  &1.00 &4.00 &12.90 &27.50 \\
                               &HOTEL    &20  &11.0  &1.00 &4.00 &14.80 &35.45 \\
                               &FLIGHT   &20  &9.20  &1.00 &4.50 &19.95 &24.25 \\ 
                               &MOVIE    &20  &8.00  &1.00 &7.00 &13.45 &34.15 \\
                               &MEDICINE &20  &6.55  &1.00 &4.20 &13.15 &19.00 \\
\midrule
\multirow{2}{*}{Multi-channel} &TRAVEL1  &12  &17.30 &1.00 &4.00 &23.10 &48.30 \\
                               &TRAVEL2  &10  &21.33 &1.00 &4.00 &23.67 &55.25 \\ 
\bottomrule
\end{tabular}
\end{table*}

\subsection{Components}
As illustrated in Equation (\ref{equ:COUT}), we formulate the COUT process in a common cyber world case, personal assistance of smart devices, in terms of three stages of thoughts: a) Thought \#1 GUI Navigation, b) Thought \#2 Underlying Reasoning, and c) Thought \#3 Personalized Recommendation. % 
Thought \#1 denotes the surface-level thought for basic embodied behavior on the device GUI, e.g., \textit{This action clicks a button named [Button\_name]}.
Thought \#2 denotes the deeper-level thought for explicit user preference, e.g., \textit{It seems the user needs items with [Restriction1], [Restriction2], and [Restriction3]}. 
Thought \#3 denotes the high-level thought for implicit user preference, e.g., \textit{I recommend [item\#1] from the item pool}.

\textbf{Challenges.} There are several key challenges to achieving COUT.
The key challenge starts with the fact that task goals are often user's ambiguous queries, lacking definitive goals matched to the observations. For example, the existing embodied agents are trained to either touch a specific object using robotic arms in 3D space or click a particular button on 2D screens. However, these golden targets are not presented in user queries as personalized preference is typically subjective and nonverbal. How to define the personalized task in embodied environments and evaluate it remains unknown and challenging.
Second, the deficiency of supportive data for COUT research poses a considerable challenge. 
The commonly used datasets are typically collected from task-oriented demonstrations. This gap highlights the need for more comprehensive datasets that can better facilitate advancements in COUT research.
Third, due to the lack of clear task definitions and the absence of supportive training environments, the methods for analyzing personalized features have not been thoroughly explored.

\section{The SmartSpot Benchmark}
\label{sec:SmartSpot}
Given the scarcity of training data for embodied agents that explicitly captures the personalized analysis highlighted in the COUT paradigm,
we propose to construct a novel benchmark named SmartSpot.

\subsection{Dataset Summary}
\label{sec:Data Collection}
We collect SmartSpot data from Meituan\footnote{\textcolor{black}{https://www.meituan.com/en-US/about-us}}, a well-known Internet service platform in China that offers a variety of life services, including food recommendations, hotel bookings, online flight ticket sales, etc. 
To create more practical personal scenes, we develop SmartSpot with two scenarios: the single-channel scenario which focuses on one type of service, and the multi-channel scenario which combines two single channels.
We select five of the most daily used single channels: Food, Hotel, Flight, Movie, and Medicine. 
For the multi-channel scenarios, we pair Flight and Food as Travel1, and Flight and Hotel as Travel2. These combinations reflect more complicated and practical situations, such as traveling to a destination and booking hotels or restaurants.
The data in every scenario are GUI action episodes which contain several steps to complete an instruction.
Specifically, each episode consists of three groups of steps, as shown in different colors in Table~\ref{tab:Datasets statistics}. The blue \colorbox{GUIcolor}{``GUI'' steps} denote a series of GUI actions, like entering and completing a search bar, to access an item pool, the yellow \colorbox{Poolcolor}{``Item Pool'' step} denotes the page for the found item pool, and the \colorbox{RScolor}{``RS'' steps} denote the details page of each item awaiting recommendations.
Each step contains a GUI screenshot, a ground-truth action (with possibly a bounding box or textual value), a list of previous actions, and an episodic instruction with a corresponding underlying requirement.  
Finally, SmartSpot covers seven scenarios that present a wealth of personal assistant tasks, supporting more embodied personalized research. The full data statistics are illustrated in Table~\ref{tab:Datasets statistics}.

\textbf{Action Space.} SmartSpot provides common human-GUI interaction operations. Following AiTW~\cite{rawles2024androidinthewild}, each action type is assigned by a \texttt{action\_type\_id} as ground truth.
The full action space is shown as follows:
% \vspace{-0.10in}
\begin{itemize}[leftmargin=*]
    \item \texttt{click(x,y)}:\texttt{4}. A click action at (x,y), where each value ranges from [0, 1], indicates the corresponding position ratio relative to the image's width and height.
    \item \texttt{type("text")}:\texttt{3}. An action that types text at (x, y). 
    
\begin{figure*}[t!]        
    \center
    {\includegraphics[width=\textwidth]{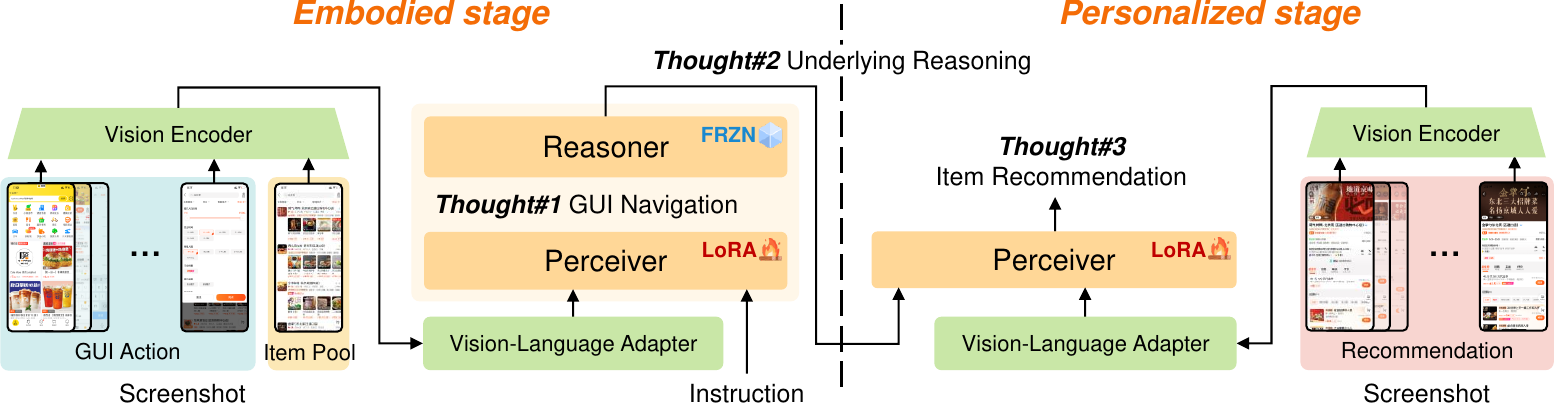}}        
    \caption{Two-stage training paradigm of SmartAgent. \textcolor{black}{Thought \#2 serves as a key intermediate result, connecting the Embodied and Personalized stages. This summarized user underlying requirement in the GUI operations greatly narrows the scope of the item pool for the next stage of personalized reasoning and item recommendation.}}
\label{fig:Two_stage_model}
\end{figure*}

    \item \texttt{scroll(direction)}: Scroll actions for the screen. Scroll up, down, left, and right are assigned with \texttt{1}, \texttt{0}, \texttt{8}, and \texttt{9}.
    \item \texttt{complete}:\texttt{10}. An action of determining if has reached the item pool.
    \item \texttt{recommendation("Yes/No")}:\texttt{2}. An action of recommending an item, as either ``Yes'' or ``No''. 
    \item \texttt{BACK}:\texttt{5}. An action for returning to the previous step.
    \item \texttt{HOME}:\texttt{6}. An action for returning to the homepage. 
\end{itemize}

\subsection{Dataset Collection \& Analysis} 
We curate SmartSpot following the real-life usage of this platform.
The process begins with generating pairs of user instructions along with the underlying requirements. 
To ensure consistency, we establish intention seeds to generate them simultaneously.
Specifically, we recruit annotators experienced on this platform to select 2-3 significant attributes as seeds for searching specific item pools for each channel.
For example, the ``[recently played]'' in Figure~\ref{fig:COUT} presents one seed result, which could be other choices that users can click on. 
By creating multiple permutations of intention seeds, we collect a diverse batch of instruction pairs while ensuring there were no duplicates.
Then, according to these instruction pairs, the annotators performed episodic GUI operations, capturing screenshots and ground-truth actions along with their bounding boxes at each step. 
All episodes are completed based on the annotators' personal usage habits on the platform.
In total, we gather 144 episodes with over 1,400 steps.

\section{SmartAgent Approach}
\label{sec:SmartAgent}
\subsection{Overview}
\label{sec:Task Overview}
Given a user instruction $i \in \mathcal{I}$ to complete, the agent will navigate through an episode comprising three groups of steps: searching for an item pool, finding the item pool, and making item recommendations. 
At time step $t$, the agent receives a screenshot observation $o_t \in \mathcal{O}$.
Then, the agent should takes an action $a_t \in \mathcal{A}$ according to its current assets $\{o_t,i,h_{t-1}\}$, where $h_{t-1}=(o_1, a_1, ...,o_{t-1}, a_{t-1})$ is the historical actions. 
The chosen action signifies either a GUI command, a signal that an item pool has been found, or a recommendation result for items within that pool. 

The primary design principles of SmartAgent are two-fold: 
i) It should process the multimodal input of high-resolution screenshot images and textual instructions, and the output of embodied action, along with textual thoughts; 
ii) It should deal with ambiguous instructions as reasoning goals at all the stages of COUT. 
We therefore transform all data of different modalities into a token sequence, illustrated below:
\begin{equation}
\label{equ:data}
\begin{split}
    &\underbrace{\text{\texttt{Generate...}}}_{\text{system message}}~\underbrace{T_{\text{image}}^{(1)},..., T_{\text{image}}^{(M)}}_{\substack{\text{screenshot image tokens}}}\underbrace{\text{\texttt{step1:...}}}_{\text{previous actions}},\\
    &\underbrace{\text{\texttt{User:I want... Agent:}}}_{\text{instruction}}~\underbrace{T_{\text{res}}^{(1)},...T_{\text{res}}^{(N)}}_{\text{response}}.
\end{split}
\end{equation}
Using this representation, we formulate the learning of SmartAgent as GPT-style auto-regressive approach, in line with~\cite{huang2023embodied}.
For instance, in Figure~\ref{fig:COUT}, given a smartphone screenshot and user's instruction ``Can you give me some music recommendations?'', we craft a query prompt as: ``Please generate the next action according to the \textless{screenshot}\textgreater
and \textless{instruction}\textgreater ''. 
Next, we will detail the training progress, evaluation, and detailed settings.

\subsection{Agent Training} 
To achieve COUT reasoning, we divide each episode into the following embodied and personalized stages to train SmartAgent successively. 
Specifically, the SmartAgent backbone LVLM functions in two roles: a Perceiver trained in our environment to predict actions or a Reasoner utilizes the original LVLMs to generate thoughts.

\textbf{Embodied Stage.}
This stage aims to complete ambiguous instructions to find the item pool. The agent takes only GUI action and item pool screenshots as visual input. 
As stated in Section~\ref{sec:Task Overview}, the multimodal assets first feed into the Perceiver to predict the specific embodied actions, referred to as Thought \#1 in COUT. 
Subsequently, the Reasoner infers step-wise action thoughts and summaries as an underlying requirement as Thought \#2. 
The Thought \#2 indicates the user's intention explicitly reflected in this stage. 
For instance, as illustrated in Figure~\ref{fig:COUT}, for user inquiries about music, Thought \#2 may include specific constraints not present in the original instruction, such as ``Justin Bieber". 
As a result, the underlying requirement serves as a key intermediate output, offering a clearer representation of user intentions for the next personalized stage.

\textbf{Personalized Stage.}
With the fine-grained underlying requirement in Thought \#2, this stage focuses on making personalized recommendations.
The same Perceiver model takes item screenshots as visual input and determines whether a recommendation is warranted, responding with either textual ``Yes'' or ``No'', which is designated as Thought \#3.

\subsection{Implementation Details}
\label{sec:training}
We choose Qwen-VL~\cite{bai2023qwen} as our backbone LVLM, which encodes visual inputs with a high resolution of 448*448.
The training of SmartAgent starts from the continual pre-training on SeeClick base model~\cite{cheng2024seeclick} for basic GUI grounding abilities. 
\textcolor{black}{Following the approach of~\cite{cheng2024seeclick}, We intuitively present numerical coordinates as natural languages without additional tokenization. 
We take 8 historical actions with screenshots during training considering the GPU memory limitation.
We train SmartAgent 15 epochs for both the embodied and personalized stages.
All baselines are trained for 15 rounds. }
Results on the ScreenSpot~\cite{cheng2024seeclick} and Mind2Web~\cite{deng2024mind2web} benchmarks are evaluated via direct testing and 10 epochs of training, as in \cite{cheng2024seeclick}.
During training, we apply LoRA~\cite{hu2021lora} to fine-tune both the visual encoder and LLM.
We utilize AdamW as the optimizer, 
starting with a learning rate of 3e-5 and a global batch size of 14. All training is conducted on two NVIDIA A100 GPUs. 

\section{Evaluation}
\label{sec:experiment}
We demonstrate SmartAgent’s capabilities by comprehensive evaluations of overall abilities in Section~\ref{sec:overall_ability} and a full spectrum of sub-tasks encompassing GUI Grounding in Section~\ref{sec:GUI_grounding}, Autonomous GUI Operation in Section~\ref{sec:autonomous_agent}, Explicit \& Implicit Personalized Reasoning in Section~\ref{sec:exp_imp_reasonging}, and Zero-shot Reasoning in Section~\ref{sec:zero_shot}. We also report more insights in Section~\ref{sec:more_insights}.

\noindent \textbf{Metrics.} 
Following most of the setting in~\cite{cheng2024seeclick}, 
we compute the cross-entropy loss for Thought \#1 and Thought \#3 with their ground-truth actions, and semantic similarity for Thought \#2 with the underlying requirement.
A bounding box may be contained in ground-truth action to verify if a click action is hit.
We therefore evaluate SmartAgent using the below metrics, in terms of embodied action metrics (following Mind2web) for Thought \#1, and personalized preference metrics for Thought \#2 and Thought \#3.
\\
\noindent \textcolor{black}{Embodied action metrics:}

\begin{itemize} 
\item \textbf{Element Accuracy (Ele.Acc)}: The accuracy of predicted coordinate with ground-truth for click and type action. 
\item \textbf{Step Successful Rate (SSR)}: The rate of steps that both the action type and value are successfully predicted.
\end{itemize}

\noindent \textcolor{black}{Personalized preference metrics:}

\begin{itemize} 
\item \textbf{Explicit Preference Accuracy (Exp.Acc)}: The semantic similarity between the predicted underlying requirement and ground truth. 
\item \textbf{Implicit Preference Accuracy (Imp.Acc)}: The accuracy of predicted item recommendations with ground truth. 
\end{itemize}

\subsection{Embodied Personalized Reasoning}
\label{sec:overall_ability}

We first investigate the comprehensive capabilities of SmartAgent in handling embodied tasks and personalized reasoning, primarily focusing on SmartSpot for validation. 
Specifically, \textcolor{black}{we categorize this experiment into simpler single-domain tasks and more complicated scenarios that mimic real multi-channel interactions.} 
We exclude the MEDICINE channel to evaluate the zero-shot ability later. We select SeeClick \textcolor{black}{(a widely used GUI agent backbone using pure visual observation), along with the well-known LLM foundation model, Qwen-VL series}, as our baselines.
\textcolor{black}{We keep the original prompt reasoning setting for all general LLM and GUI Specialist baselines to show their features and deficiencies on the COUT reasoning task. 
Thus, only embodied action results are reported.
} 
% \vspace{-1cm}

\begin{figure*}[t!]        
    \center
    {\includegraphics[width=\textwidth]{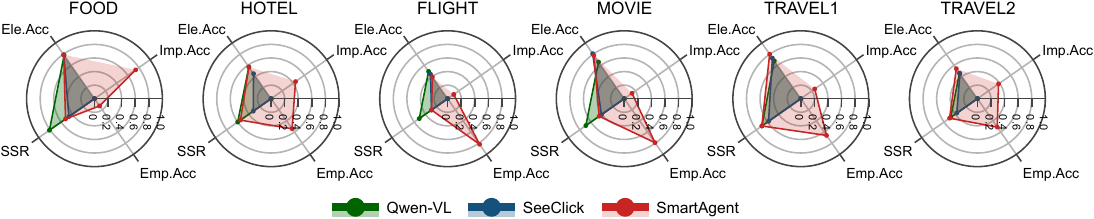}}     
    \caption{\textbf{Comparison of methods on SmartSpot.} SmartAgent performs comparable and even better with GUI specialist model and general LLM in all scenarios. Notably, in the more complex scenarios TRAVEL1 and TRAVEL2, which involve longer episodes, SmartAgent consistently shows excellent performance across all embodied and personalized metrics.}
\label{fig:MAIN_result_channel}
\end{figure*}

\noindent \textbf{Results \& analysis.} Table~\ref{tab:main_result} shows SmartAgent's average performance across all channels. It achieved the best results in the Ele.Acc metric and ranked second in SSR, indicating strong foundational embodied perception ability. \textcolor{black}{This comprehensive understanding of GUI operations supports the prediction of} 71\% of users' underlying requirements, afterward leading to a 24\% accuracy in item recommendations.
More importantly, Figure~\ref{fig:MAIN_result_channel} shows that \textcolor{black}{SmartAgent consistently performs the best in all multi-channel scenarios, which highlights its superior reasoning capabilities in real-world applications compared to specialized GUI agents and general LLM that lack deeper personalized cognition. }
\textcolor{black}{Qwen2-VL had unexpectedly poor results that probably were caused by highly limited tunable LoRA parameters compared to Qwen-VL under the same train epochs. 
Qwen2-VL will not be discussed further and the transferability of COUT across different backbone architectures warrants future investigation.}
Overall, these evaluations confirm SmartAgent's comprehensive capabilities on both embodied and personalized tasks.

\begin{table}[h]
\caption{Comparison of methods on SmartSpot. 
}
\label{tab:main_result}
\centering
\begin{tabular}{p{1.7cm}<{\centering} p{1.1cm}<{\centering} p{1.0cm}<{\centering} p{1.2cm}<{\centering} p{1.2cm}<{\centering}}
\toprule
\multirow{2}{*}{Methods} &\multicolumn{4}{c}{SmartSpot} \\ 
\cmidrule(r){2-5}   
           &Ele.Acc           &SSR           &Exp.Acc           &Imp.Acc  \\ 
\midrule
Qwen2-VL   &0.04          &0.03          &/             &/  \\
Qwen-VL    &0.61          &\textbf{0.64} &/             &/  \\
SeeClick   &0.61          &0.40          &/             &/  \\ 
\midrule
SmartAgent &\textbf{0.64} &0.50          &\textbf{0.71} &\textbf{0.24}  \\
\bottomrule
\end{tabular}
\end{table}

\vspace{0.10in}
\subsection{GUI Grounding}
\label{sec:GUI_grounding}
An important consideration is whether training with personalized capabilities leads to catastrophic forgetting of pre-trained embodied abilities. In this section, we evaluate SmartAgent on a renowned GUI Grounding benchmark ScreenSpot~\cite{cheng2024seeclick} to assess its foundational perception of raw GUI data. The comparative baselines are divided into two main categories: GUI specialist models and general LLMs.

\noindent \textbf{Results \& analysis.} As shown in Table~\ref{app:tab:screenspot}, SmartAgent achieves second-best results in most metrics, even securing the top position in Desktop-Icon/Widget. This indicates that training with personalized capabilities not only preserves foundational embodied abilities but also enhances proficiency in operations involving user intent. This is also the primary achievement of the COUT paradigm. 
\textcolor{black}{Note that we do not overly require SmartAgent to maintain a SOTA level on such traditional test environment that do not include personalized factors. Our primary goal in this section is to ensure that it retains sufficient basic capabilities in tasks after incorporating personalized considerations.}

\vspace{0.10in}
\subsection{Autonomous GUI Operation}
\label{sec:autonomous_agent}

\begin{figure}[t!]        
    \center
    {\includegraphics[width=0.95\columnwidth]{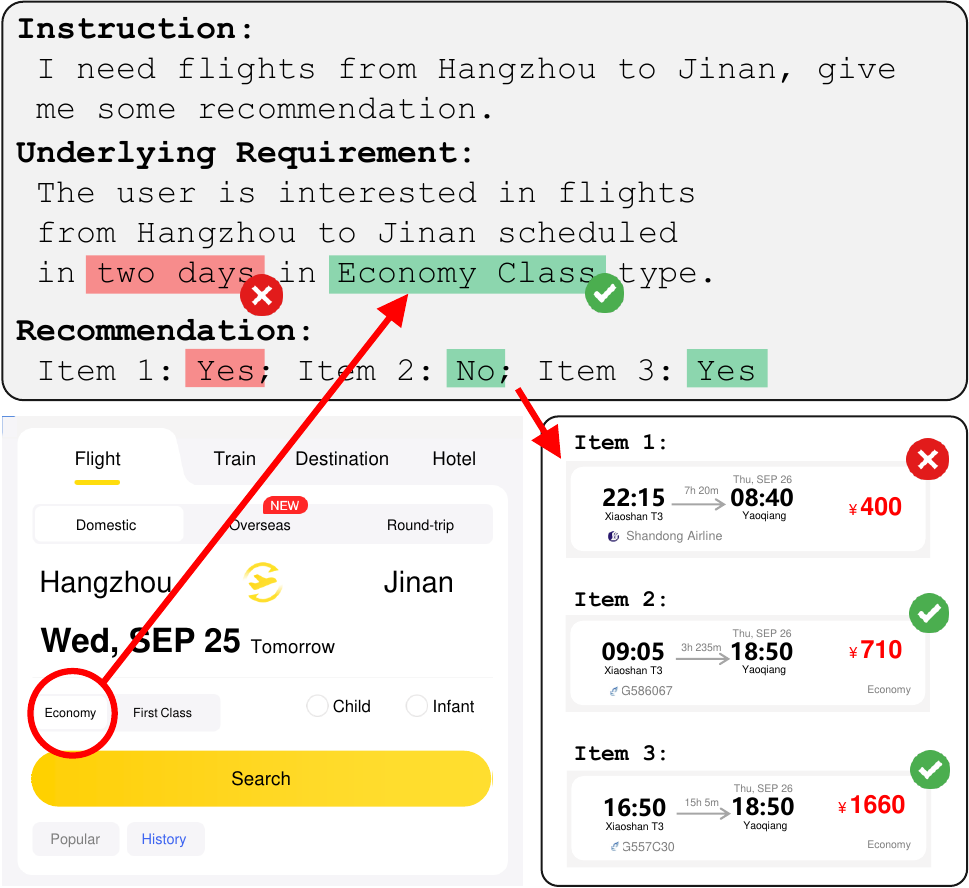}}
    \caption{Case study for embodied personalized reasoning.} 
% \vskip -1.0in
\label{fig:Case_study}
\end{figure}
Automated execution of human instructions is a foundational capability of embodied agents. In this section, we validate SmartAgent's basic abilities to handle GUI action episodes autonomously using the classic GUI agent benchmark Mind2Web~\cite{deng2024mind2web}. 

\noindent \textbf{Results \& analysis.} We transfer the well-trained SmartAgent on SmartSpot to Mind2Web. As shown in Appendix Table~\ref{app:tab:mind2web}, it achieves second place generally in the pure vision-based baselines across three downstream scenarios. This indicates that SmartAgent possesses strong embodied generalization capabilities, enabling seamless integration into various downstream tasks.
\textcolor{black}{Similarly, we do not expect SmartAgent to surpass the SOTA model in this basic task. The first-tier autonomous agent capabilities that have not been forgotten in the backbone model have laid the foundation for subsequent personalized reasoning tasks.}

\vspace{0.10in}
\subsection{Explicit \& Implicit Personalized Reasoning}
\label{sec:exp_imp_reasonging}

In this section, we present a case study to demonstrate SmartAgent's performance in explicit user underlying reasoning and implicit item recommendation tasks. As shown in Figure~\ref{fig:Case_study}, SmartAgent correctly predicted the action of clicking on the economy class option in the visual observation. This prediction is reflected in the summarization of the user's underlying requirements. Leveraging this textual representation of explicit needs, SmartAgent made recommendations for the last two flight options, effectively addressing the user's implicit requirements.

\begin{table*}[t!]
\centering
\caption{GUI grounding results on ScreenSpot.}
% \vskip 0.10in
\label{app:tab:screenspot}
\begin{tabular}{p{2.0cm}<{\centering} p{1.4cm}<{\centering} c p{1.9cm}<{\centering} c p{1.9cm}<{\centering} c p{1.9cm}<{\centering} p{1.5cm}<{\centering}}
\toprule
\multirow{2.5}{*}{LVLMs} &\multirow{2.5}{*}{\makecell[c]{GUI\\Specific}} &\multicolumn{2}{c}{Mobile} &\multicolumn{2}{c}{Desktop} &\multicolumn{2}{c}{Web} &\multirow{2}{*}{Average} \\ 
\cmidrule(r){3-4} \cmidrule(r){5-6} \cmidrule(r){7-8}
& & Text & Icon/Widget & Text  & Icon/Widget  & Text  & Icon/Widget &   \\ 
\midrule
GPT-4V     &\textcolor{darkred}{\ding{55}} &22.6\%  &24.5\% &20.2\% &11.8\% &9.2\%   &8.8\%  &16.2\% \\
MiniGPT-v2 &\textcolor{darkred}{\ding{55}} &8.4\%   &6.6\%  &6.2\%  &2.9\%  &6.5\%   &3.4\%  &5.7\%  \\
Qwen-VL    &\textcolor{darkred}{\ding{55}} &9.5\%   &4.8\%  &5.7\%  &5.0\%  &3.5\%   &2.4\%  &5.2\%  \\ 
Fuyu       &\textcolor{darkgreen}{\ding{51}} &41.0\%  &1.3\%  &33.0\% &3.6\%  & 33.9\% &4.4\%  &19.5\%  \\
CogAgent   &\textcolor{darkgreen}{\ding{51}} &67.0\%  &24.0\% &\textbf{74.2\%} &20.0\% &\textbf{70.4\%} &\underline{28.6\%} & 47.4\% \\
SeeClick   &\textcolor{darkgreen}{\ding{51}} &\textbf{78.0\%} &\textbf{52.0\%} &72.2\% &30.0\% &55.7\% & \textbf{32.5\%} &\textbf{53.4\%}  \\ 
\midrule
SmartAgent &\textcolor{darkgreen}{\ding{51}} &\underline{77.3\%} &\underline{49.8\%} &\underline{72.7\%} &\textbf{32.1\%} &\underline{57.4\%} &24.8\% &\underline{52.4\%}  \\ 
\bottomrule
\end{tabular}
\end{table*}

\begin{table}[t!]
\caption{Results of zero-shot reasoning on channel MEDICINE. 
}
\label{tab:zero_shot}
\centering
\begin{tabular}{p{1.6cm}<{\centering} p{1.2cm}<{\centering} p{1.0cm}<{\centering} p{1.2cm}<{\centering} p{1.2cm}<{\centering}}
\toprule
\multirow{2.5}{*}{Type} &\multicolumn{4}{c}{MEDICINE} \\   
\cmidrule(r){2-5}
           &Ele.Acc  &SSR  &Exp.Acc  &Imp.Acc  \\ 
\midrule
Zero-shot  &\cellcolor[rgb]{0.9, 0.9, 0.9}0.04 &\cellcolor[rgb]{0.9, 0.9, 0.9}0.08 &\cellcolor[RGB]{130,182,130}\textbf{0.77} &\cellcolor[RGB]{174,214,174}0.14  \\
\midrule
Full-stage &\cellcolor[RGB]{214,234,214}0.64 &\cellcolor[RGB]{214,234,214}0.50 &\cellcolor[RGB]{214,234,214}0.71 &\cellcolor[RGB]{214,234,214}0.24  \\
\bottomrule
\end{tabular}
\end{table}

\begin{table}[t!]
\caption{Results of end-to-end training and two-stage training. 
}
\label{tab:more_insights}
\centering
\begin{tabular}{p{1.8cm}<{\centering} p{1.1cm}<{\centering} p{1.0cm}<{\centering} p{1.1cm}<{\centering} p{1.2cm}<{\centering}}
\toprule
        &Ele.Acc       &SSR           &Exp.Acc       &Imp.Acc  \\ 
\midrule
Two-stage &0.64          &0.50          &\textbf{0.71} &0.24  \\
End-to-End &\textbf{0.67} &\textbf{0.53} &0.66          &\textbf{0.31}  \\
\bottomrule
\end{tabular}
\end{table}

\subsection{Zero-shot Reasoning} 
\label{sec:zero_shot}

Zero-shot perception is the ultimate goal for embodied agents, enabling them truly to learn from interactions with their environment. In this section, we utilize the MEDICINE channel in SmartSpot to evaluate SmartAgent's zero-shot performance in unseen scenarios.

\noindent \textbf{Results \& analysis.} As shown in Table~\ref{tab:zero_shot}, SmartAgent surprisingly exceeds its average performance achieved through fine-tuning on SmartSpot in the Exp.Acc metric. 
\textcolor{black}{We argue that in cold-start scenarios like MEDICINE, the training data may be very limited, rendering fine-tuning ineffective or even leading to overfitting. SmartAgent's zero-shot capability can compensate for this limitation by leveraging general GUI knowledge to interact directly with new scenarios. This enhancement is also evident in the subsequent item recommendation task, which approaches the performance of full-stage fine-tuning.}
Overall, this demonstrates SmartAgent's robustness in interpreting users' explicit intentions and also indicates a preliminary zero-shot reasoning capability.

\subsection{More Insights into COUT}
\label{sec:more_insights}

In this section, we provide deeper insights into COUT reasoning, focusing on the two-stage training and end-to-end training settings (without underlying thought generation). As shown in Table~\ref{tab:more_insights}, SmartAgent with end-to-end training exhibits even slightly better embodied performance, although it is limited in user underlying perceptions. 
\textcolor{black}{This may be due to the large number of long action episodes, which forces the model to overfit underlying requirements or even accumulate serious errors, which then negatively affects the subsequent personalized stages.}
We, therefore, argue that balancing embodied perceptions in response to changing environments with reliable personal service is crucial for future COUT research.

\section{Conclusion and Future Work}
\label{sec:conclusion}
In this paper, we introduce a novel embodied reasoning paradigm, COUT, which for the first time defines an embodied personalized task. 
We establish a clear definition and components of the COUT paradigm and analyze its challenges.
To address these challenges, we propose SmartAgent to instantiate COUT through a two-stage training from essential GUI reasoning to high-level user thought reasoning.
To benchmark this progress in SmartAgent, we created SmartSpot, the first embodied AI benchmark featuring explicit personalization evaluations.
Results on SmartSpot demonstrate the effectiveness and proficiency of SmartAgent over full-stage embodied personalized reasoning tasks.
Furthermore, SmartAgent showcases the key capability of zero-shot embodied reasoning, highlighting its potential for efficient adaptation in new scenarios.

As for the future work, we plan to further extend the scale of the benchmark, with more extensive experiments and analysis. We also plan to apply the proposed method to the real user devices.

{
    \small
    \bibliographystyle{ieeenat_fullname}
    \bibliography{main}
}

\clearpage
\onecolumn

\section{More Results}
\label{app:sec:more_results}
This section shows SmartAgent's performance results on autonomous GUI operation on the benchmark Mind2Web in Table~\ref{app:tab:mind2web}.
% \clearpage

\begin{table}[!h]
\centering
\caption{Results of Autonomous GUI operation on Mind2Web.}
% \vskip 0.10in
\label{app:tab:mind2web}
\begin{tabular}{p{2.2cm}<{\centering} p{1.2cm}<{\centering} p{1.1cm}<{\centering} c p{1.1cm}<{\centering} p{1.1cm}<{\centering} c p{1.1cm}<{\centering} p{1.1cm}<{\centering} c p{1.1cm}<{\centering}}
\toprule
\multirow{2.5}{*}{Methods} & \multirow{2.5}{*}{\makecell[c]{Pure\\Visual}} & \multicolumn{3}{c}{Cross-Task} & \multicolumn{3}{c}{Cross-Website} & \multicolumn{3}{c}{Cross-Domain} \\ 
\cmidrule(r){3-5} \cmidrule(r){6-8} \cmidrule(r){9-11}
& & Ele.Acc & Op.F1 & SSR & Ele.Acc & Op.F1  & SSR & Ele.Acc & Op.F1 & SSR   \\ 
\midrule
MindAct (gen) &\textcolor{darkred}{\ding{55}} &20.2 &52.0 &17.5 &13.9 &44.7 &11.0 &14.2 &44.7 &11.9  \\
MindAct       &\textcolor{darkred}{\ding{55}} &5.1  &75.7 &52.0 &42.0 &65.2 &38.9 &42.1 &66.5 &39.6  \\
GPT-3.5-Turbo &\textcolor{darkred}{\ding{55}} &20.3 &56.6 &17.4 &19.3 &48.8 &16.2 &21.6 &52.8 &18.6  \\
GPT-4         &\textcolor{darkred}{\ding{55}} &41.6 &60.6 &36.2 &35.8 &51.1 &30.1 &37.1 &46.5 &26.4  \\ 
\midrule
Qwen-VL      &\textcolor{darkgreen}{\ding{51}}&15.9 &\underline{86.7} &13.3  &13.2 &\textbf{83.5} &9.2 &14.1 &\underline{84.3} & 12.0  \\
SeeClick     &\textcolor{darkgreen}{\ding{51}}&\textbf{28.3} &\textbf{87.0}&\textbf{25.5} &\textbf{21.4} &\underline{80.6} &\textbf{16.4} &\textbf{23.2} &\textbf{84.8} &\textbf{20.8}  \\ 
\midrule
SmartAgent   &\textcolor{darkgreen}{\ding{51}}&\underline{24.5}&80.2 &\underline{20.6} &\underline{18.9} &74.9 &\underline{15.0} &\underline{20.3} &77.9 &\underline{17.1}  \\
\bottomrule
\end{tabular}
\end{table}

\end{document}